\documentclass{bmvc2k}

\usepackage{amssymb}
\usepackage{extarrows}
\usepackage{booktabs}
\usepackage{multirow}
\usepackage{color}
\usepackage{colortbl}
\usepackage{bm}
\usepackage{url}
\usepackage{threeparttable}
\usepackage{graphicx}
\usepackage{makecell}
\usepackage{appendix}

\usepackage{algorithm}
\usepackage{algorithmic}

\title{A Strong Baseline for Semi-Supervised Incremental Few-Shot Learning}

\addauthor{Linglan Zhao\textsuperscript{$\ast$}}{llzhao@sjtu.edu.cn}{1}
\addauthor{Dashan Guo\textsuperscript{2}, Yunlu Xu\textsuperscript{$\dag$}\textsuperscript{2}, Liang Qiao}{{guodashan,xuyunlu,qiaoliang6}@hikvision.com}{2}
\addauthor{Zhanzhan Cheng\textsuperscript{2}, Shiliang Pu\textsuperscript{2}, Yi Niu}{{chengzhanzhan,pushiliang,niuyi}@hikvision.com}{2}
\addauthor{Xiangzhong Fang}{xzfang@sjtu.edu.cn}{1}

\addinstitution{
 Dept. of Electronic Engineering\\
 Shanghai Jiao Tong University\\
 Shanghai, China
}
\addinstitution{
 Hikvision Research Institute\\
 Hangzhou, China
}

\runninghead{L. ZHAO ET AL.}{A Strong Baseline for Semi-Supervised Incremental FSL}

\begin{document}

\maketitle

\begin{abstract}
Few-shot learning (FSL) aims to learn models that generalize to novel classes with limited training samples.
Recent works advance FSL towards a scenario where unlabeled examples are also available and propose semi-supervised FSL methods.
Another line of methods also cares about the performance of base classes in addition to the novel ones and thus establishes the incremental FSL scenario.
In this paper, we generalize the above two under a more realistic yet complex setting,
named by Semi-Supervised Incremental Few-Shot Learning (S$^2$I-FSL).
To tackle the task, we propose a novel paradigm containing two parts:
(1) a well-designed meta-training algorithm for mitigating ambiguity between base and novel classes caused by unreliable pseudo labels and
(2) a model adaptation mechanism to learn discriminative features for novel classes while preserving base knowledge using few labeled and all the unlabeled data.
Extensive experiments on standard FSL, semi-supervised FSL, incremental FSL, and the firstly built S$^2$I-FSL benchmarks demonstrate the effectiveness of our proposed method.
\end{abstract}

\vspace{-0.4cm}
\section{Introduction}
\label{sec:intro}
\vspace{-0.3cm}
Despite the great success of deep learning models in recent years, modern deep learning approaches rely heavily on abundant labeled data which can be prohibitively expensive to acquire. In contrast, humans can quickly learn novel concepts with only a few samples. To tackle this problem,~\emph{Few-Shot Learning (FSL)}~\cite{MatchingNet,ProtoNet,Relationnet,MAML} methods are proposed to bridge the gap between supervised learning and the intriguing property of human cognition.

\begin{figure}
\centering
\vspace{-0.1cm}
\includegraphics[width=0.99\textwidth]{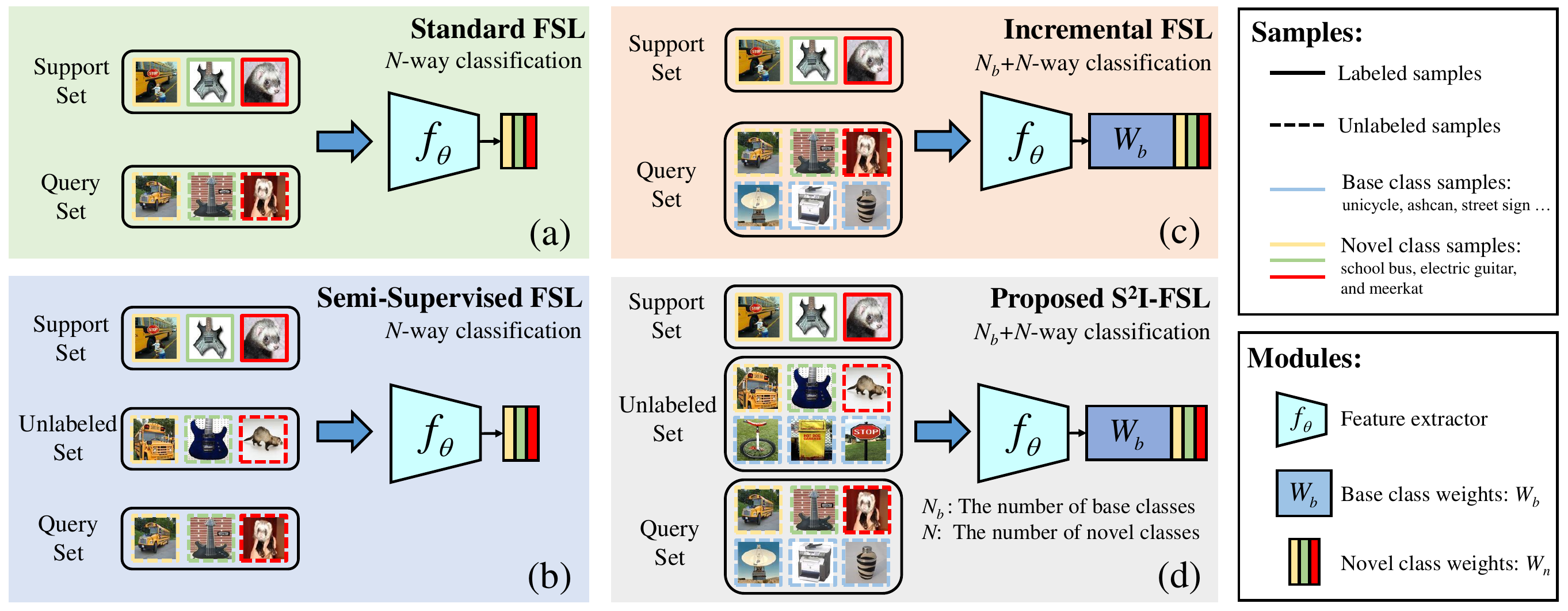}
\vspace{-0.3cm}
\caption{Comparisons between our proposed S$^2$I-FSL and other FSL benchmarks. $N_b$ and $N$ are the numbers of base and novel classes, respectively. Best viewed in color.} \label{setting}
\vspace{-1em}
\end{figure}

The \textit{standard FSL} setting is shown in Fig.~\ref{setting} (a) that given new tasks of limited supervision, the system turns the prior knowledge from base tasks.
Recent works~\cite{Soft_KMeans,Transmatch,TAFSSL,PTN} advance the few-shot paradigm towards a scenario where unlabeled examples are available, namely the \emph{semi-supervised FSL} shown in Fig.~\ref{setting} (b).
Exploring extra unlabeled data along with the limited supervised ones makes sense since unlabeled data is much easier to obtain than the labeled.
However, the standard or semi-supervised FSL focuses solely on recognizing few novel classes, while ignoring the ability to handle the previously learned base classes. 
To prevent forgetting prior knowledge for base tasks, \emph{incremental FSL}~\cite{Dynamic,Imprinting,Attractor,XtarNet} is built for simultaneously regarding performance on base and novel classes, which shows some realistic significance, as shown in Fig.~\ref{setting} (c).
Nevertheless, there is still room for improvement as only supervised data is used which is hard to adapt to new classes.

From a more realistic perspective, besides few annotated samples, we expect to additionally utilize more easily available unlabeled data containing both base and novel class samples to overcome the performance limitation, especially the performance of novel classes.
Therefore, we propose a new benchmark incorporating the above semi-supervised and incremental settings together called \emph{Semi-Supervised Incremental Few-Shot Learning (S$^2$I-FSL)}, which generalizes previous assumptions under a more realistic yet complex setting (Fig.\ref{setting} (d)).

Intuitively, a straightforward solution for the proposed S$^2$I-FSL is to simply combine the existing semi-supervised and incremental FSL approaches, but it does not work well.
For detail, as illustrated in Section \ref{subsec:meta-inc-baseline}, we utilize a simple yet efficient meta-learning-based method as our incremental FSL baseline (denoted as Meta-Inc-Baseline).
When extended with unlabeled data in a semi-supervised manner by the mainstreaming methods including weights refinement~\cite{Soft_KMeans,TAFSSL,BD-CSPN}, label propagation~\cite{TPN,EPNet} and FixMatch~\cite{FixMatch}, unfortunately, it can not perform well.
We find that novel class weights refined with incorrectly predicted pseudo-labels will overlap with base class weights and lead to ambiguity among classes in the embedding space.
For example, the weight of novel class ``school bus'' in Fig.~\ref{setting} (d) is likely to be incorrectly refined with unlabeled samples from base class ``unicycle'' (both have the \emph{wheel} pattern), ``ashcan'' (similar color and texture), and `` street sign'' (two objects often appear in the same image), especially when only a few labeled ``school bus'' samples are given as in few-shot scenarios.
Overall, S$^2$I-FSL brings additional challenges in the FSL field:
(1) 
using unlabeled data from both base and novel classes leads easily to ambiguity among classes in the embedding space, showing performance degradation;
(2) unbalanced sample number of different classes in unlabeled set
instead of a balanced number in each novel class makes it harder to learn the classifier.
In this paper, we mainly handle the former challenge and show the comparison effect on the latter one.

Upon the above, we propose a novel S$^2$I-FSL paradigm of two phases including two key components, a novel meta-training scheme and a well-designed model adaptation mechanism.
In the meta-training phase, the training set is sampled using a mechanism to mimic the semi-supervised scenario during testing. As a result, the model learns to make use of unlabeled data for incremental learning and mitigates confusion between base and novel classes.
Considering that the feature extractor is trained on the training set, it may not contain discriminative information for recognizing novel classes unseen during training, 
a model adaptation scheme is proposed. Concretely, besides standard classification loss on few labeled data, a contrastive learning loss function is applied to unlabeled data to explore the novel class distribution implicitly.
In addition, we also use distillation technique to preserve the classification ability on base classes.

To summarize, the contributions of this work are three-fold: (1) A Semi-Supervised Incremental Few-Shot Learning (S$^2$I-FSL) benchmark is proposed to generalize previous semi-supervised FSL and incremental FSL under a more realistic and challenging setting; (2) Technically, we propose an efficient meta-training paradigm and a model adaptation mechanism for the newly built task, which can be regarded as a baseline for future researches; (3) Extensive experiments on standard FSL, incremental FSL, semi-supervised FSL, and the S$^2$I-FSL benchmark demonstrate the effectiveness of our proposed method.
\begin{figure}
\centering
\vspace{-0.1cm}
\includegraphics[width=0.99\textwidth]{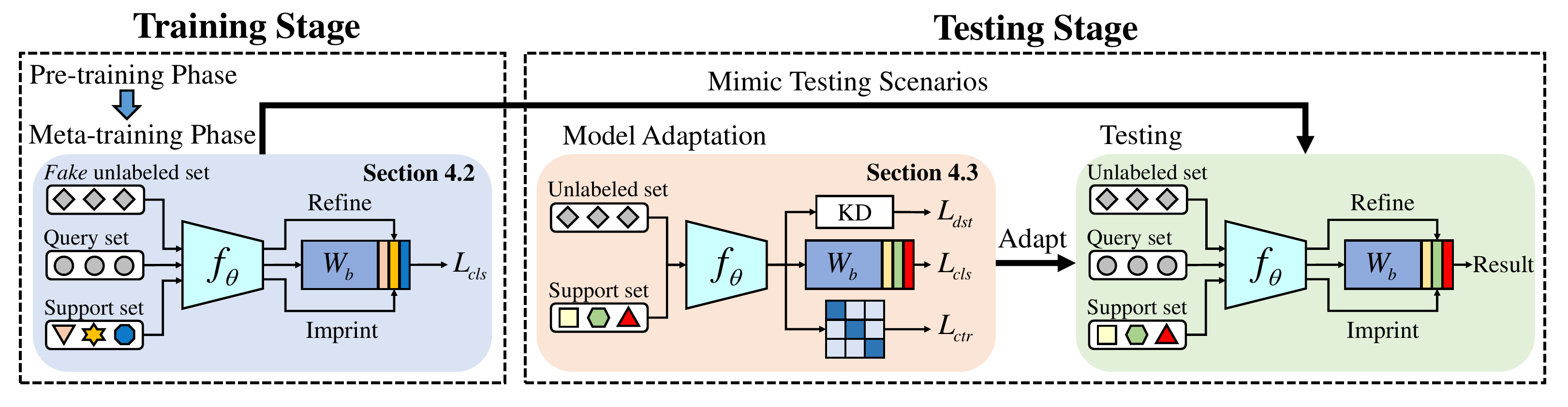}
\vspace{-0.3cm}
\caption{An overview of our proposed method for S$^2$I-FSL. The left and right parts illustrate the model training and testing stages, respectively. Best viewed in color.} \label{schematic}
\vspace{-1em}
\end{figure}
\vspace{-0.3cm}

\section{Related works}
\label{sec:relatedwork}
\vspace{-0.3cm}
\subsection{Few-Shot Learning}
\vspace{-0.2cm}
Few-shot learning (FSL) aims to learn novel categories with limited labeled examples. Previous episodic-training~\cite{MatchingNet} based FSL works can be roughly divided into 4 categories. \emph{Metric learning based} approaches~\cite{MatchingNet,ProtoNet,Relationnet} aim at learning the similarity metric between query (\textit{i.e.}, test) samples and support (\textit{i.e.}, training) samples. The goal of \emph{initialization based} FSL methods~\cite{MAML,MAML++,LEO} is to learn good model initialization for fast adaptation to current tasks. \emph{Weight generation based} methods~\cite{Dynamic,Imprinting,Act2Param} directly predict classification weights for novel classes. Moreover, \emph{hallucination based} approaches~\cite{Imaginary,Saliency-guided,AFHN} learn a generator for data augmentation. However, recent works~\cite{CloserLook,Boosting,RFS,Meta-baseline} reveal that a well-trained feature extractor, rather than other complex meta-trained modules, is the key component to achieve state-of-the-art performance.
This motivates us to adopt a modified version of~\cite{Meta-baseline} as a baseline method (Meta-Inc-Baseline) for incremental FSL and S$^2$I-FSL benchmark.

\vspace{-0.3cm}
\subsection{Few-Shot Incremental Learning}
\vspace{-0.2cm}
Incremental FSL involves training a model to classify novel classes while preserving the capability of handling previously learned classes.
There are two main different settings in the literature. The first setting is also called generalized (or dynamic) FSL, which contains two sequences of tasks. The first task contains abundant training samples from base classes, while the second task contains only a few novel class training samples. Methods proposed in this setting include novel class weight generation~\cite{Dynamic,Imprinting,RelationalGF,LCwoF}, attention-based regularization~\cite{Attractor} and task adaptive features extraction~\cite{XtarNet,TapNet}.
The second setting~\cite{TOPIC,LVQ,FSLL,CEC} shares the same core idea with the first one but considers several sequences of tasks instead of two. 
We build S$^2$I-FSL mainly based on the first setting since extra unlabeled data are not available in the second setting. In addition, the second setting uses unique data splits which makes it harder to compare to other standard/transductive/semi-supervised FSL works.

\vspace{-0.3cm}
\subsection{Semi-Supervised Few-Shot Learning}
\vspace{-0.2cm}
Semi-supervised/transductive FSL works take additional unlabeled data into account to boost the performance of FSL. \emph{Pseudo-labeling based} methods and \emph{graph-based} methods are two main lines of efforts. Pseudo-labeling based methods consist of prototype refinement~\cite{Soft_KMeans,TAFSSL,BD-CSPN,PT+MAP}, self-training~\cite{LST,ICI} and entropy minimization~\cite{Transductive_ft,TIM}. As for graph-based approaches~\cite{TPN,DPGN,EPNet,PTN}, graph models are constructed to propagate information from labeled data to unlabeled data. Among these approaches, we mainly focus on the utilization of prototype refinement for its effectiveness and extensiveness.

\vspace{-0.3cm}
\section{Task Formulation}
\label{sec:problem formulation}
\vspace{-0.2cm}
In this section, we introduce the detailed settings for the proposed S$^2$I-FSL benchmark.

\textbf{Dataset Splits.} For evaluating classification performance on both base and novel classes, we follow the dataset splits in incremental FSL~\cite{Dynamic,Attractor,XtarNet}.
Concretely, $\mathcal{D}_{\text{base}}$ and $\mathcal{D}_{\text{novel}}$ are two datasets for base classes $\mathcal{C}_{\text{base}}$ and novel classes $\mathcal{C}_{\text{novel}}$ respectively, and $\mathcal{C}_{\text{base}} \cap \mathcal{C}_{\text{novel}}=\varnothing$.
$\mathcal{D}_{\text{base}}$ is comprised of $\mathcal{D}_{\text{base/train}}$, $\mathcal{D}_{\text{base/val}}$ and $\mathcal{D}_{\text{base/test}}$ as base class training, validation and testing sets ($\left| \mathcal{C}_{\text{base}} \right| = N_b$). $\mathcal{D}_{\text{novel}}$ consists of three splits of disjoint novel classes: $\mathcal{D}_{\text{novel/train}}$, $\mathcal{D}_{\text{novel/val}}$ and $\mathcal{D}_{\text{novel/test}}$ for meta-training, validation and testing, respectively.

\textbf{Training Stage.} As shown in the left part of Fig.~\ref{schematic}, the training stage is split into \emph{pre-training phase} and \emph{meta-training phase}. In the pre-training phase, feature extractor $f_{\theta}$ and the base class weights $W_{b}$ are trained on $\mathcal{D}_{\text{base/train}}$ as in standard supervised learning. After that, in the meta-training phase, the model is further trained using both base class training set $\mathcal{D}_{\text{base/train}}$ and novel class training set $\mathcal{D}_{\text{novel/train}}$ in an episodic training manner~\cite{MatchingNet}. 

\textbf{Testing Stage.} Following the evaluation protocol in~\cite{Dynamic,Attractor,XtarNet}, average accuracy on sampled incremental FSL episodes (\textit{i.e.}, tasks) is reported.
For each test episode, $N$ novel categories are randomly chosen from $\mathcal{D}_{\text{novel/test}}$ ($\mathcal{D}_{\text{novel/val}}$ for validation), which are relabeled with new labels from $N_b + 1$ to $N_b + N$.
After that, $K$ per-class training samples are sampled from these $N$ classes (\emph{$N$-way $K$-shot}) to construct a support set $\mathcal{S}$.
The query set $\mathcal{Q}$ consists of samples from both novel and base classes,
\textit{i.e.}, additional examples from the selected $N$ novel classes and base class test examples from $\mathcal{D}_{\text{base/test}}$ ($\mathcal{D}_{\text{base/val}}$ for validation).
The main difference between S$^2$I-FSL and incremental FSL is that unlabeled data is available in S$^2$I-FSL. Two different settings are considered: a \emph{transductive} setting where the unlabeled set $\mathcal{U}$ is exactly the query set $\mathcal{Q}$; a \emph{semi-supervised} setting where $\mathcal{U}$ contains extra unlabeled samples from both base classes and novel classes.

\vspace{-0.3cm}
\section{Methodology}
\label{sec:methodology}
\vspace{-0.3cm}
\subsection{Preliminary}
\label{subsec:meta-inc-baseline}
\vspace{-0.2cm}
Since Meta-Baseline \cite{Meta-baseline} provides an effective yet strong baseline for standard FSL, we extend it for incremental setting, noted as Meta-Inc-Baseline.
During the \emph{pre-training phase}, the feature extractor $f_{\theta}(\mathbf{x}) \in \mathbb{R}^{d}$ is trained using a cosine classifier on $\mathcal{D}_{\text{base/train}}$ as in standard supervised training. The conditional probability of sample $\mathbf{x}$ belonging to class $k$ is:
\begin{equation}
P\left(y=k \mid \mathbf{x}; W_b\right)=\frac{\exp \left(\gamma \cos\left(f_{\theta}\left(\mathbf{x}\right), w_{k}\right)\right)}{\sum_{j} \exp \left(\gamma \cos\left(f_{\theta}\left(\mathbf{x}\right), w_{j}\right)\right)}, \label{cosine classifier}
\end{equation}
where $W_{b}=\left[w_{1}, \ldots, w_{N_{b}}\right] \in \mathbb{R}^{d \times N_{b}}$ is the base class weights and $\gamma$ is a learnable scalar.

In the \emph{testing stage}, novel class weights can be constructed as class means of the support set $\mathcal{S}$, \textit{i.e.}, the novel class weights $W_{n}= \left[p_{1}, \ldots, p_{N}\right] \in \mathbb{R}^{d \times N}$ are comprised of prototypes~\cite{ProtoNet} of each category where $\mathcal{S}_{j} = \{(\mathbf{x}, y) \mid (\mathbf{x}, y) \in \mathcal{S}, y = N_b + j\}$:
\vspace{-0.2cm}
\begin{equation}
p_{j}=\frac{1}{\left|\mathcal{S}_{j}\right|} \sum_{\left(\mathbf{x}, y\right) \in \mathcal{S}_{j}} f_{\theta}\left(\mathbf{x}\right). \label{prototypes}
\vspace{-0.2cm}
\end{equation}
For extending Meta-Baseline to handle both base and novel classes, the overall classification weights $W$ is constructed by concatenating the base class weights $W_b$ and novel class weights $W_n$, \textit{i.e.} $W = \left[ W_b, W_n \right] \in \mathbb{R}^{d \times (N_b+N)}$. Finally, the model makes predictions on query set $\mathcal{Q}$ containing both base and novel class samples based on Eq.~\ref{cosine classifier} with $W_b$ replaced with $W$.

After pre-training is a \emph{meta-training phase}.
Since the model is evaluated on incremental FSL tasks, instead of only optimizing $f_{\theta}$ and discarding $W_{b}$~\cite{Meta-baseline}, we meta-train both $f_{\theta}$ and $W_{b}$ by sampling incremental FSL tasks from $\mathcal{D}_{\text{base/train}}$ and $\mathcal{D}_{\text{novel/train}}$ as in Algorithm~\ref{algorithm:Meta-Inc-Baseline}.


\begin{algorithm}[!htb]
\footnotesize
\caption{Meta-training algorithm for Meta-Inc-Baseline.}
\label{algorithm:Meta-Inc-Baseline}
\begin{algorithmic}[1]
\REQUIRE pre-trained $f_{\theta}$ and $W_b$, learning rate for $\theta$ and $W_b$ respectively: $\eta_1$ and $\eta_2$.
\ENSURE Meta-trained feature extractor $f_{\theta}$ and base class weights $W_b$.
\WHILE {not done}
    \STATE $\left\{\mathcal{S}, \mathcal{Q}_{n}\right\} \leftarrow$ Sample support and query set from $\mathcal{D}_{\text{novel/train}}$
    \STATE $\left\{\mathcal{Q}_{b}\right\} \leftarrow$ Sample query set from $\mathcal{D}_{\text{base/train}}$
    \STATE $\mathcal{Q} = \mathcal{Q}_{b} \cup \mathcal{Q}_{n}$ \textcolor{blue}{$//$ Construct an incremental FSL episode} 
    \STATE Construct overall classification weight $W = \left[ W_b, W_n \right]$ using Eq.~\ref{prototypes} \label{construct_W_using_Eq2}
    \STATE Get predictions $\hat{Y}_{q} \in \mathbb{R}_{+}^{n_q \times (N_{b}+N)}$ on $\mathcal{Q}$ ($n_q = \left| \mathcal{Q} \right|$) using Eq.~\ref{cosine classifier} with $W_b \leftarrow W$ \label{make_predictions1}
    \STATE Calculate cross-entropy loss $\mathcal{L}_{cls}\left(\hat{Y}_{q}, Y_q\right)$ based on $\hat{Y}_{q}$ and ground truth labels $Y_q$
    \STATE Update: $\theta \leftarrow \theta - \eta_1 \nabla_{\theta} \mathcal{L}_{cls}$, $W_b \leftarrow W_b - \eta_2 \nabla_{W_b} \mathcal{L}_{cls}$
\ENDWHILE
\end{algorithmic}
\end{algorithm}
\vspace{-1em}

\vspace{-0.3cm}
\subsection{Prototype Refinement with Fake Unlabeled Data}
\label{subsec:meta-training paradigm}
\vspace{-0.2cm}
Upon Meta-Inc-Baseline, we further generalize to S$^2$I-FSL in Section~\ref{subsec:meta-training paradigm} and Section~\ref{subsec:model adaptation}. Since prototype refinement using pseudo labels has been proved effective in semi-supervised methods~\cite{Soft_KMeans,TAFSSL,BD-CSPN,PT+MAP}, an intuitive solution to extend Meta-Inc-Baseline to S$^2$I-FSL is to refine the novel class weights $W_n$ with unlabeled data. Meanwhile, base class weights $W_b$ trained on abundant base class samples remain unchanged.

Concretely, predictions $\hat{Y}_{u} \in \mathbb{R}_{+}^{n_u \times (N_{b}+N)}$ on $\mathcal{U}$ ($n_u = \left| \mathcal{U} \right|$) can be obtained as in line~\ref{construct_W_using_Eq2}-\ref{make_predictions1} of Algorithm~\ref{algorithm:Meta-Inc-Baseline} similar to $\hat{Y}_{q}$. Since $\mathcal{U}$ contains unlabeled samples from both base and novel classes, samples predicted as base classes should be filtered out as they are viewed as distractors in the refinement process: $\hat{Y}^{\prime} = \hat{Y}_{u} [:, N_b:N_b+N] \in \mathbb{R}_{+}^{n_u \times N}$ where $[:]$ denotes slicing operation. Based on $\hat{Y}^{\prime}$, novel class prototypes $\{ p_{j}^{\prime} \}_{j=1}^{N}$ can be re-estimated :
\vspace{-0.2cm}
\begin{equation}
p_{j}^{\prime} = \frac{\sum_{i=1}^{n_u} \hat{Y}^{\prime}_{ij} f_{\theta}(\mathbf{x}_i) + \sum_{(\mathbf{x}_k, y_k) \in \mathcal{S}_j} f_{\theta}(\mathbf{x}_k)}{\sum_{i=1}^{n_u} \hat{Y}^{\prime}_{ij} + \left| \mathcal{S}_j \right|}. \label{prototype re-estimate}
\vspace{-0.2cm}
\end{equation}
Then we progressively refine $W_n= \left[p_{1}, \ldots, p_{N}\right] \in \mathbb{R}^{d \times N}$:
\vspace{-0.2cm}
\begin{equation}
p_{j} \leftarrow \alpha \cdot p_{j}^{\prime} + (1 - \alpha) \cdot p_{j}, \label{prototype refine}
\vspace{-0.2cm}
\end{equation}
where $\alpha \in \left(0, 1\right]$. After iterating the above process of Eq.~\ref{prototype re-estimate}-\ref{prototype refine} for $n_{\text{steps}}$ loops, query samples are evaluated with $W = \left[ W_b, W_n \right]$ where $W_n$ is the refined novel class weights.

However, as discussed in Section~\ref{sec:intro} and verified in Section~\ref{sec:experiments}, this straightforward modification leads to little improvement or even obvious degradation. It is because that novel class weights refined with incorrectly predicted pseudo-labels will be confused with similar base classes. For utilization of unlabeled samples and mitigating ambiguity, a novel meta-training algorithm with \emph{fake}\footnote{We use the terminology \emph{fake} unlabeled since a subset is randomly sampled from the supervised training set in each meta-training episode, but its labels are not used to simulate the testing scenario.} unlabeled examples is proposed in Algorithm~\ref{algorithm:pseudo-unlabeled} as a substitution for Algorithm~\ref{algorithm:Meta-Inc-Baseline}. Concretely, compared to labeled support set $\mathcal{S}$ or query set $\mathcal{Q}$ whose labels are used in the loss function, the label information of sampled unlabeled set $\mathcal{U}$ in each training episode is not used to mimic the testing scenario. By default, $n_{\text{steps}} = 1$, $\alpha=1.0$ in training. 


\begin{algorithm}[!htb]
\footnotesize
\caption{Meta-training with \emph{fake} unlabeled data.}
\label{algorithm:pseudo-unlabeled}
\begin{algorithmic}[1]
\REQUIRE pre-trained $f_{\theta}$ and $W_b$, learning rate for $\theta$ and $W_b$: $\eta_1$ and $\eta_2$. refinement parameters: $n_{\text{steps}} = 1$, $\alpha=1.0$.
\ENSURE Meta-trained feature extractor $f_{\theta}$ and base class weights $W_b$.
\WHILE {not done}
    \STATE $\left\{\mathcal{S}, \mathcal{Q}_{n}, \mathcal{U}_{n} \right\} \leftarrow$ Sample support, query and unlabeled set from $\mathcal{D}_{\text{novel/train}}$
    \STATE $\left\{\mathcal{Q}_{b}, \mathcal{U}_{b} \right\} \leftarrow$ Sample query and unlabeled set from $\mathcal{D}_{\text{base/train}}$
    \STATE $\mathcal{Q} = \mathcal{Q}_{b} \cup \mathcal{Q}_{n}$, $\mathcal{U} = \mathcal{U}_{b} \cup \mathcal{U}_{n}$ \textcolor{blue}{$//$ Construct an incremental FSL episode with \emph{fake} unlabeled set} 
    \STATE Get refined $W_n$ using Eq.~\ref{prototype re-estimate}-\ref{prototype refine} based on $\mathcal{U}$, and construct classification weight $W = \left[ W_b, W_n \right]$
    \STATE Get predictions $\hat{Y}_{q} \in \mathbb{R}_{+}^{n_q \times (N_{b}+N)}$ on $\mathcal{Q}$ ($n_q = \left| \mathcal{Q} \right|$) using Eq.~\ref{cosine classifier} with $W_b \leftarrow W$
    \STATE Calculate cross-entropy loss $\mathcal{L}_{cls}\left(\hat{Y}_{q}, Y_q\right)$ based on $\hat{Y}_{q}$ and ground truth labels $Y_q$
    \STATE Update: $\theta \leftarrow \theta - \eta_1 \nabla_{\theta} \mathcal{L}_{cls}$, $W_b \leftarrow W_b - \eta_2 \nabla_{W_b} \mathcal{L}_{cls}$
\ENDWHILE
\end{algorithmic}
\end{algorithm}
\vspace{-1em}

\vspace{-0.3cm}
\subsection{Model Adaptation without Forgetting}
\vspace{-0.2cm}
\label{subsec:model adaptation}
As described in Section~\ref{sec:intro}, feature extractor $f_{\theta}$ trained on the training set may not contain discriminative information for novel classes during testing~\cite{CloserLook,crossdomainfewshot,PTN}. However, as observed in few-shot incremental literature~\cite{Dynamic,Attractor,XtarNet}, directly finetuning models on few labeled novel class samples will cause not only overfitting to the novel class but also catastrophic forgetting of the old classes. Fortunately, since additional unlabeled data is available, $\{f_{\theta}, W_{b}\}$ can be further adapted to the current test episode $\mathcal{T} = \left\{\mathcal{S}, \mathcal{Q}, \mathcal{U} \right\}$ in the testing stage.

As shown in the right part of Fig.~\ref{schematic}, for learning better representations for novel classes, we explore both labeled support set $\mathcal{S}$ using cross-entropy loss $\mathcal{L}_{cls}$ for classification,
and the unlabeled set $\mathcal{U}$ using a contrastive loss~\cite{SimCLR} $\mathcal{L}_{ctr}$:
\vspace{-0.3cm}
\begin{equation}
\mathcal{L}_{ctr} = -\frac{1}{2B}\sum_{i, j=1}^{B} \log \frac{\exp \left(\cos \left(f_{i}, f_{j}\right) / \tau_1 \right)}{\sum_{k \neq i} \exp \left(\cos \left(f_{i}, f_{k}\right) / \tau_1 \right)}, \label{contrastive loss}
\vspace{-0.3cm}
\end{equation}
where $B$ is the batch size to sample batches from $\mathcal{U}$, $\tau_1$ is a temperature parameter and $(f_{i}, f_{j})$ is a positive feature pair (different augmentations from the same image) based on $f_{\theta}$.

For preserving the knowledge learned from base classes, we adopt a distillation~\cite{distilling} loss $\mathcal{L}_{dst}$ on base class predictions using unlabeled data $\mathcal{U}$:
\vspace{-0.3cm}
\begin{equation}
\mathcal{L}_{dst} = -\frac{1}{B} \sum_{i=1}^{B} \sum_{k=1}^{N_b} \bar{z}_{ik} \log z_{ik}, \label{distillation loss}
\vspace{-0.2cm}
\end{equation}
where $\bar{z}_{ik}$ and $z_{ik}$ are $\tau_2$ softened predictions of the $i$-th sample belonging to class $k$ from previous meta-trained parameters $\{ \bar{\theta},\bar{W}_b \}$ and current adapted ones $\{ \theta,W_b \}$, respectively:
\vspace{-0.3cm}
\begin{equation}
\bar{z}_{ik} = \frac{\exp \left(\gamma \cos\left(f_{\bar{\theta}}\left(\mathbf{x}_{i}\right), \bar{w}_{k}\right) / \tau_2 \right)}{\sum_{j=1}^{N_b} \exp \left(\gamma \cos\left(f_{\bar{\theta}}\left(\mathbf{x}_{i}\right), \bar{w}_{j}\right) / \tau_2 \right)}, \  {z}_{ik} = \frac{\exp \left(\gamma \cos\left(f_{{\theta}}\left(\mathbf{x}_{i}\right), {w}_{k}\right) / \tau_2 \right)}{\sum_{j=1}^{N_b} \exp \left(\gamma \cos\left(f_{{\theta}}\left(\mathbf{x}_{i}\right), {w}_{j}\right) / \tau_2 \right)}. \label{logits}
\vspace{-0.2cm}
\end{equation}

The overall loss function is formulated as: $\mathcal{L} =  w_{cls} \cdot \mathcal{L}_{cls} + w_{ctr} \cdot \mathcal{L}_{ctr} + w_{dst} \cdot \mathcal{L}_{dst}$ where $w_{cls}$, $w_{ctr}$ and $w_{dst}$ (hyperparameters in Appendix C) control the tradeoff between each term.
\vspace{-1em}

\section{Experiments}
\label{sec:experiments}
\vspace{-0.2cm}
\subsection{Experiment Setup}
\label{subsec:Experiments Setup}
\vspace{-0.2cm}
\textbf{Datasets.} We follow the common incremental FSL setting~\cite{Dynamic,Attractor,XtarNet}, and conduct experiments on \emph{mini}-ImageNet~\cite{MatchingNet} and \emph{tiered}-ImageNet~\cite{Soft_KMeans}
with $64$ and $200$\footnote{Due to the dataset splits in incremental FSL~\cite{Attractor,XtarNet}, the number of base classes is $200$ instead of $351$.} base classes, respectively.
More details of dataset statistics can be found in Appendix A.

\textbf{Evaluation Protocol.} The performance is evaluated using the average accuracy and corresponding $95\%$ confidence interval over $600$ randomly sampled episodes with a \emph{fixed seed}, which makes results more stable and reliable. 
In incremental FSL and S$^2$I-FSL, we report joint accuracy \text{Acc$_{\text{all/all}}$} (\text{Acc.} for short) which is the accuracy of classifying all the test samples to all (both base and novel) classes. To measure the performance gap between joint accuracy and individual accuracy, following \cite{Attractor,XtarNet}, metric $\Delta$ is also reported. Concretely, $\Delta = \left( \Delta_{\text{b}} + \Delta_{\text{n}} \right) / 2$ is computed as the mean of performance degradation on base and novel classes.
$\Delta_{\text{b}}$ ($\Delta_{\text{n}}$) is the performance degradation between the accuracy of classifying base (novel) class test samples to all classes \text{Acc$_{\text{b/all}}$} (\text{Acc$_{\text{n/all}}$}) and the accuracy of classifying base (novel) class test samples to only base (novel) classes \text{Acc$_{\text{b/b}}$} (\text{Acc$_{\text{n/n}}$}):
\vspace{-0.2cm}
\begin{equation}
\Delta_{\text{b}} = \text{Acc$_{\text{b/all}}$} - \text{Acc$_{\text{b/b}}$}, \  \Delta_{\text{n}} = \text{Acc$_{\text{n/all}}$} - \text{Acc$_{\text{n/n}}$}. \label{delta base and novel}
\vspace{-0.2cm}
\end{equation}

\textbf{Implementation Details.} In most of the experiments, ResNet12~\cite{ResNet,TADAM} (detailed in Appendix B) is adopted as the backbone $f_{\theta}$ to have fair comparisons between other
methods. Unless specified in S$^2$I-FSL, unlabeled base and novel class samples are maintained in equal proportion. Concretely, in each test episode, we sample $5$ novel classes with $1$/$5$ per-class training samples ($5$-way $1$/$5$-shot) for support set $\mathcal{S}$. Query set $\mathcal{Q}$ contains $5$$\times$$15$ novel class test samples and $75$ examples sampled uniformly from the base class test set~\cite{Attractor,XtarNet}. After adopting the semi-supervised FSL~\cite{LST,ICI,TAFSSL} setting, unlabeled set $\mathcal{U}$ in S$^2$I-FSL contains $5\times30$ / $5\times50$ novel class samples and $150$ / $250$ base class samples for $1$/$5$-shot setting.
We leave more details of hyper-parameter settings in Appendix C.
\vspace{-0.2cm}

\subsection{Experimental Results}
\label{subsec:Experimental Results}
\vspace{-0.2cm}
\textbf{Incremental FSL.} We first conduct experiments to evaluate the robustness of the proposed Meta-Inc-Baseline on the incremental FSL classification task.  The upper parts of Table~\ref{tab:Incremental FSL mini-imagenet} and Table~\ref{tab:Incremental FSL tiered-imagenet} show the results (denoted as \emph{Inductive}) on the \emph{mini}-ImageNet and \emph{tiered}-ImageNet, respectively.
Compared to other approaches, our Meta-Inc-Baseline is able to achieve much higher joint accuracy and diminish the gap between joint accuracy and individual accuracy.
The direct performance enhancement is mainly attributed to the extension of the simple but strong baseline of Meta-baseline~\cite{Meta-baseline}. The detailed verification is described in Appendix D. 
Moreover, the effectiveness of Meta-Inc-Baseline on incremental FSL benchmark with multi sessions~\cite{TOPIC} is also verified in Appendix E.

\textbf{S$^2$I-FSL.}
The proposed method is then generalized into two settings where unlabeled data is available: the \emph{Transductive} setting where unlabeled set $\mathcal{U}$ is exactly the query set $\mathcal{Q}$, and the \emph{Semi-supervised} setting where extra unlabeled set $\mathcal{U}$ is available. 
Besides our proposed method, extensions with prevailing semi-supervised FSL methods such as label propagation~\cite{TPN} (a graph model is constructed with base and novel class weights as labeled vertices and unlabeled samples as unlabeled vertices), prototype refinement~\cite{TAFSSL} and semi-supervised learning approach FixMatch~\cite{FixMatch} are also evaluated for comparison. 
\begin{table}[!htb]
    \scriptsize
    \begin{center}
    \begin{threeparttable}
		\begin{tabular}{l l c || c c || c c}
			\toprule[1pt]
			\multirow{2}*{\textbf{Method}} & \multirow{2}*{\normalsize$f_{\theta}$} & \multirow{2}*{\textbf{Type}} & \multicolumn{2}{c}{\textbf{64+5-way 1-shot}} & \multicolumn{2}{c}{\textbf{64+5-way 5-shot}} \\ \cmidrule{4-7}
			~ & ~ & ~ & \textbf{Acc.} & \textbf{$\Delta$} & \textbf{Acc.} & \textbf{$\Delta$} \\
			\midrule
            Imprint~\cite{Imprinting} & Res10 & \multirow{9}*{\makecell[c]{Incremental FSL\\(Inductive)}} & 41.34$\pm$0.54 & -23.79 & 46.34$\pm$0.54 & -25.25 \\
            LwoF~\cite{Dynamic}  & Res10 & ~ & 49.65$\pm$0.64 & -14.47 & 59.66$\pm$0.55 & -12.35 \\
            Attractor~\cite{Attractor} & Res10 & ~  & 54.95$\pm$0.30 & -11.84 & 63.04$\pm$0.30 & -10.66 \\
            TapNet~\cite{TapNet} & Res12$^{\ddagger}$ & ~ & 54.38$\pm$0.59 & -12.88 & 64.02$\pm$0.51 & -10.98 \\
            XtarNet~\cite{XtarNet} & Res12$^{\ddagger}$ & ~ & 54.96$\pm$0.61 & -13.36 & 64.88$\pm$0.47 & -10.41 \\
            Meta-Inc-Baseline (ours)  & Res12$^{\ddagger}$ & ~ & 59.27$\pm$0.43 & -11.30 & 69.74$\pm$0.36 & -9.90 \\
            LCwoF \emph{unlim}~\cite{LCwoF} & Res12 & ~ & 58.16$\pm$N/A & - N/A & 66.88$\pm$N/A & - N/A \\
            Meta-Inc-Baseline (ours)  & Res12 & ~ & \textbf{60.65$\pm$0.42} & \textbf{-11.23} & \textbf{71.52$\pm$0.35} & \textbf{-9.30} \\
            \midrule \midrule
            Meta-Inc-Baseline + Graph~\cite{TPN} & Res12 & \multirow{4}*{\makecell[c]{S$^2$I-FSL\\(Transductive)}} & 55.02$\pm$0.41 & -18.32 & 65.71$\pm$0.38 & -15.57 \\
            Meta-Inc-Baseline + FixMatch~\cite{FixMatch} & Res12 & ~ & 57.14$\pm$0.46 & -13.24 & 72.07$\pm$0.36 & -9.33 \\
			Meta-Inc-Baseline + PR~\cite{TAFSSL} & Res12 & ~ & 62.74$\pm$0.54 & -13.57 & 72.60$\pm$0.37 & -9.50 \\
            Our proposed method  & Res12 & ~ & \textbf{68.43$\pm$0.54} & \textbf{-8.31} & \textbf{74.97$\pm$0.35} & \textbf{-7.95} \\
            \midrule
            Meta-Inc-Baseline + Graph~\cite{TPN} & Res12 & \multirow{4}*{\makecell[c]{S$^2$I-FSL\\(Semi-supervised)}} & 50.41$\pm$0.36 & -23.53 & 62.82$\pm$0.38 & -18.04 \\
            Meta-Inc-Baseline + FixMatch~\cite{FixMatch} & Res12 & ~ & 56.60$\pm$0.46 & -13.59 & 72.00$\pm$0.38 & -9.49 \\
			Meta-Inc-Baseline + PR~\cite{TAFSSL} & Res12 & ~ & 64.53$\pm$0.54 & -12.46 & 73.48$\pm$0.35 & -9.00 \\
            Our proposed method & Res12 & ~ & \textbf{70.33$\pm$0.51} & \textbf{-8.07} & \textbf{75.91$\pm$0.33} & \textbf{-7.79} \\
			\bottomrule[1pt]
		\end{tabular}
	
	 \begin{tablenotes}
        \scriptsize
        \item[]$^{\ddagger}$: ResNet12 backbone with $256$ output channels as~\cite{XtarNet,TapNet}. PR: directly Prototype Refinement using unlabeled data.
      \end{tablenotes}
    \end{threeparttable}

	\end{center}
	\vspace{-1em}
	\caption{Incremental FSL and S$^2$I-FSL results on \emph{mini}-ImageNet.} 
    \label{tab:Incremental FSL mini-imagenet}
	\vspace{-0.5cm}
\end{table}

As shown in the bottom parts of Table~\ref{tab:Incremental FSL mini-imagenet} and Table~\ref{tab:Incremental FSL tiered-imagenet}, label propagation~\cite{TPN} is not suitable for the more challenging task since it always results in lower accuracy and degrades $\Delta$ metric. While FixMatch~\cite{FixMatch} can get about $0.5\%$ performance gain in 5-shot case, the accuracy decreases significantly in 1-shot case. It is because that FixMatch relies heavily on the quality of pseudo-labels that cannot be obtained in few-shot scenarios especially the 1-shot case. Moreover, the direct combination of strong baseline and prototype refinement~\cite{TAFSSL} can get a limited improvement on \emph{mini}-ImageNet but leads to obvious performance degradation on \emph{tiered}-ImageNet. Since \emph{tiered}-ImageNet contains more base classes, confusion between base and novel classes caused by unreliable pseudo labels is prone to happen.

In contrast, our proposed method can obtain significant performance gains on both joint accuracy and resistance to forgetting consistently.
In \emph{Transductive} setting, compared to Meta-Inc-Baseline + PR, we get $5.7\%$/$12.0\%$ 1-shot classification performance gains and $5.3\%$/$10.1\%$ $\Delta$ metric improvements on the two datasets, respectively.
It is because that our proposed method can alleviate confusion between base and novel classes, and is well adapted to the current test episode to learn discriminative features for novel classes.
In \emph{Semi-Supervised} setting, our method outperforms Meta-Inc-Baseline by about $10\%$ in 1-shot classification on both benchmarks, which shows the superiority of utilizing the unlabeled set.

For better understanding the challenges of S$^2$I-FSL, we report detailed results in Table~\ref{tab:challenges of S2I-FSL.}. Although all the methods can improve \text{Acc$_{\text{n/n}}$} metric which is the only focus of standard/transductive/semi-supervised FSL methods, other competitors obtain decreased joint accuracy or only marginal overall improvement. This indicates that S$^2$I-FSL is a more difficult task and poses challenges that have not been solved by existing works. 
We emphasize that there is a trade-off between \text{Acc$_{\text{b/all}}$} and \text{Acc$_{\text{n/all}}$}, and balancing between them is significant for the final result. 
For example, although \text{Acc$_{\text{b/all}}$} increases using graph models~\cite{TPN}, \text{Acc$_{\text{n/all}}$} decreases dramatically which results in degraded joint accuracy. As mentioned in Section~\ref{sec:intro}, there is an imbalance between base and novel classes. Since the number of base classes is larger than that of novel classes, label information from base vertices ($N_b=200$) is more likely propagated to unlabeled vertices which decreases \text{Acc$_{\text{n/all}}$}. 
The marginal improvement of FixMatch~\cite{FixMatch} in 5-shot case is attributed to that the increase in \text{Acc$_{\text{n/all}}$} slightly outweighs the decrease in \text{Acc$_{\text{b/all}}$}. 
However, this limited improvement is unstable, since it gets degraded performance in more challenging 1-shot scenarios on both datasets.
As for prototype refinement~\cite{TAFSSL}, \text{Acc$_{\text{b/all}}$} degrades significantly due to the confusion between base and novel classes. Concretely, some base class unlabeled samples are incorrectly predicted as novel classes and are used to refine novel weights. Base class testing samples may be predicted as novel if the corresponding weight is refined with the similar but incorrect base class unlabeled samples. For example, as discussed in Section~\ref{sec:intro}, novel class weight ``school bus'' in Fig.~\ref{setting} (d) is likely to be incorrectly refined with unlabeled samples from base class ``unicycle'' (same \emph{wheel} pattern), ``ashcan'' (similar color and texture), and `` street sign'' (often appear in the same image) which causes confusion.
By contrast, our method can boost both \text{Acc$_{\text{b/all}}$} and \text{Acc$_{\text{n/all}}$}, which accounts for the obvious improvement on joint accuracy.
\vspace{-0.3cm}

\begin{table}[!htb]
    \scriptsize
	\begin{center}
	\begin{threeparttable}
		\begin{tabular}{l l c || c c || c c}
			\toprule[1pt]
			\multirow{2}*{\textbf{Method}} & \multirow{2}*{\normalsize$f_{\theta}$} & \multirow{2}*{\textbf{Type}} & \multicolumn{2}{c}{\textbf{200+5-way 1-shot}} & \multicolumn{2}{c}{\textbf{200+5-way 5-shot}} \\ \cmidrule{4-7}
			~ & ~ & ~ & \textbf{Acc.} & \textbf{$\Delta$} & \textbf{Acc.} & \textbf{$\Delta$} \\
			\midrule
            Imprint~\cite{Imprinting} & Res18 & \multirow{7}*{\makecell[c]{Incremental FSL\\(Inductive)}} & 40.83$\pm$0.45 & -22.29 & 53.87$\pm$0.48 & -17.18 \\
            LwoF~\cite{Dynamic}  & Res18 & ~ & 53.42$\pm$0.56 & -9.59 & 63.22$\pm$0.52 & -7.27 \\
            Attractor~\cite{Attractor} & Res18 & ~  & 56.11$\pm$0.33 & \textbf{-6.11} & 65.52$\pm$0.31 & \textbf{-4.48} \\
            TapNet~\cite{TapNet} & Res18 & ~ & 55.42$\pm$0.59 & -8.09 & 65.38$\pm$0.53 & -6.37 \\
            XtarNet~\cite{XtarNet} & Res18 & ~ & 56.57$\pm$0.60 & -8.02 & 66.52$\pm$0.49 & -6.31 \\
            Meta-Inc-Baseline (ours)  & Res18 & ~ & \textbf{63.16$\pm$0.51} & -7.53 & \textbf{73.28$\pm$0.40} & -5.35 \\
            LCwoF \emph{unlim}~\cite{LCwoF} & Res12 & ~ & 58.82$\pm$N/A & -N/A & 66.61$\pm$N/A & -N/A \\
            Meta-Inc-Baseline (ours)  & Res12 & ~ & 61.92$\pm$0.50 & -7.87 & 72.98$\pm$0.38 & -5.39 \\
            \midrule \midrule
            Meta-Inc-Baseline + Graph~\cite{TPN} & Res12 & \multirow{4}*{\makecell[c]{S$^2$I-FSL\\(Transductive)}} & 59.73$\pm$0.56 & -11.25 & 70.18$\pm$0.41 & -8.58 \\
            Meta-Inc-Baseline + FixMatch~\cite{FixMatch} & Res12 & ~ & 57.49$\pm$0.59 & -8.12 & 73.34$\pm$0.38 & -5.46 \\
			Meta-Inc-Baseline + PR~\cite{TAFSSL} & Res12 & ~ & 57.99$\pm$0.61 & -15.58 & 72.08$\pm$0.39 & -7.47 \\
            Our proposed method & Res12 & ~ & \textbf{70.03$\pm$0.57} & \textbf{-5.41} & \textbf{75.73$\pm$0.37} & \textbf{-4.53} \\
            \midrule
            Meta-Inc-Baseline + Graph~\cite{TPN} & Res12 & \multirow{4}*{\makecell[c]{S$^2$I-FSL\\(Semi-supervised)}} & 51.87$\pm$0.48 & -19.71 & 65.86$\pm$0.41 & -12.63 \\
            Meta-Inc-Baseline + FixMatch~\cite{FixMatch} & Res12 & ~ & 56.56$\pm$0.59 & -8.37 & 73.22$\pm$0.39 & -5.57 \\
			Meta-Inc-Baseline + PR~\cite{TAFSSL} & Res12 & ~ & 59.35$\pm$0.58 & -14.55 & 72.68$\pm$0.39 & -6.93 \\
            Our proposed method  & Res12 & ~ & \textbf{71.64$\pm$0.56} & \textbf{-4.97} & \textbf{76.24$\pm$0.36} & \textbf{-4.21} \\
			\bottomrule[1pt]
		\end{tabular}
	
    \end{threeparttable}
	
	\end{center}
	\vspace{-1em}
	\caption{Incremental FSL and S$^2$I-FSL results on \emph{tiered}-ImageNet.} 
	\label{tab:Incremental FSL tiered-imagenet}
	\vspace{-0.3cm}
\end{table}

\vspace{-0.2cm}
\begin{table}[!htb]
    \scriptsize
	\begin{center}
	\begin{threeparttable}
		\begin{tabular}{l || l r@{.}l l l l l}
			\toprule[1pt]
			\multirow{2}*{\textbf{\makecell[l]{Method\\(Semi-supervised)}}} & \multicolumn{7}{c}{\textbf{200+5-way 5-shot}} \\  \cmidrule{2-8}
			~ & \textbf{Acc.} & \multicolumn{2}{c}{\textbf{$\Delta$}} & \textbf{Acc$_{\text{b/all}}$} & \textbf{Acc$_{\text{n/all}}$} & \textbf{Acc$_{\text{b/b}}$} & \textbf{Acc$_{\text{n/n}}$} \\ \midrule
            Meta-Inc-Baseline (baseline)                    & 72.98 & -5&39  & 65.91 & 80.05 & 73.36 & 83.39 \\ \midrule
            Meta-Inc-Baseline + Graph~\cite{TPN}            & 65.86 {\color{red}$\downdownarrows$} & -12&63 {\color{red}$\downdownarrows$} & 70.65 {\color{green}$\upuparrows$} & 61.07 {\color{red}$\downdownarrows$} & 71.28 {\color{red}$\downarrow$} & 85.70 {\color{green}$\upuparrows$} \\
            Meta-Inc-Baseline + FixMatch~\cite{FixMatch}    & 73.22 {\color{green}$\uparrow$} & -5&57  & 64.13 {\color{red}$\downarrow$} & 82.30 {\color{green}$\upuparrows$} & 73.10 & 84.47 {\color{green}$\uparrow$} \\
            Meta-Inc-Baseline + PR~\cite{TAFSSL}            & 72.68 {\color{red}$\downarrow$} & -6&93 {\color{red}$\downarrow$}  & 60.32 {\color{red}$\downdownarrows$} & 85.04 {\color{green}$\upuparrows$} & 73.36 & 85.86 {\color{green}$\upuparrows$} \\ 
            Our proposed method                             & \textbf{76.24} {\color{green}$\upuparrows$} & \textbf{-4}&\textbf{21} {\color{green}$\uparrow$}  & 67.58 {\color{green}$\upuparrows$} & 84.90 {\color{green}$\upuparrows$} & 73.08 & \textbf{87.82} {\color{green}$\upuparrows$} \\ 
			\bottomrule[1pt]
		\end{tabular}

	\begin{tablenotes}
        \scriptsize
        \item[]{\color{green}$\uparrow$}({\color{green}$\upuparrows$}): Certain metric increases (significantly). {\color{red}$\downarrow$}({\color{red}$\downdownarrows$}): Certain metric decreases (significantly).
      \end{tablenotes}
    \end{threeparttable}
	
	\end{center}
	\vspace{-1em}
	\caption{Detailed evaluations of S$^2$I-FSL on \emph{tiered}-ImageNet 5-shot setting.}
	\label{tab:challenges of S2I-FSL.}
	\vspace{-0.3cm}
\end{table}

\textbf{Semi-Supervised/Transductive FSL.} Our model trained for S$^2$I-FSL is also directly applied to regular semi-supervised and transductive FSL benchmarks. Here, we only need to perform classification on novel classes where unlabeled data solely contains novel class samples. As shown in Table~\ref{tab:SSFSL mini-imagenet}, our method consistently outperforms other approaches using the same or deeper backbones. It can be attributed to the proposed Algorithm~\ref{algorithm:pseudo-unlabeled} for effectively utilizing unlabeled samples and the model adaptation mechanism for learning discriminative features of novel classes. Experiments on \emph{tiered}-ImageNet are provided in Appendix F.
\vspace{-0.3cm}

\begin{table}[!htb]
    \scriptsize
	\begin{center}
	\begin{threeparttable}
		\begin{tabular}{r c c c || r c c c}
			\toprule[1pt]
			\multirow{2}*{\textbf{Method}} & \multirow{2}*{\normalsize$f_{\theta}$} &  \multicolumn{2}{c||}{\textbf{Transductive FSL}} & \multirow{2}*{\textbf{Method}} & \multirow{2}*{\normalsize$f_{\theta}$} & \multicolumn{2}{c}{\textbf{Semi-supervised FSL}}\\
			~ & ~ & \textbf{1-shot} & \textbf{5-shot} & ~ & ~ & \textbf{1-shot} & \textbf{5-shot} \\
			\midrule
            TPN\cite{TPN}                & Conv4 & 55.51$\pm$0.86 & 69.86$\pm$0.65 & Soft k-M\cite{Soft_KMeans}     & Conv4 & 50.41$\pm$0.31 & 64.39$\pm$0.24 \\
            TEAM\cite{TEAM}              & Res12 & 60.07$\pm$N/A  & 75.90$\pm$N/A  & LST\cite{LST}                  & Res12 & 70.10$\pm$1.90 & 78.70$\pm$0.80 \\
            BD-CSPN\cite{BD-CSPN}        & Res12 & 65.94$\pm$N/A  & 79.23$\pm$N/A  & LR+ICI\cite{ICI}               & Res12 & 71.11$\pm$1.15 & 81.25$\pm$0.69 \\
            LR+ICI\cite{ICI}             & Res12 & 68.70$\pm$1.10 & 79.69$\pm$0.66 & EPNet\cite{EPNet}              & Res12 & 75.36$\pm$1.01 & 84.07$\pm$0.60 \\
            DPGN\cite{DPGN}	             & Res12 & 67.77$\pm$0.32 & 84.60$\pm$0.43 & TransMatch\cite{Transmatch}    & WRN28 & 63.02$\pm$1.07 & 82.24$\pm$0.59 \\
            Completion\cite{Completion}  & Res12 & 73.13$\pm$0.85 & 82.06$\pm$0.54 & ICA+MSP\cite{TAFSSL}           & DenseNet & 78.55$\pm$0.25 & 85.41$\pm$0.13 \\
			\midrule
			Our proposed                  & Res12 & \textbf{77.37$\pm$0.85} & \textbf{86.02$\pm$0.41} & Our proposed  & Res12  & \textbf{79.12$\pm$0.83} & \textbf{87.16$\pm$0.38} \\
			\bottomrule[1pt]
		\end{tabular}
		
    \end{threeparttable}

	\end{center}
	\vspace{-1em}
	\caption{Regular semi-supervised/transductive FSL results on \emph{mini}-ImageNet.}
	\label{tab:SSFSL mini-imagenet}
	\vspace{-0.2cm}
\end{table}

\vspace{-0.2cm}
\subsection{Ablation Studies}
\label{subsec:Ablation Studies}
\vspace{-0.2cm}
\textbf{Effectiveness of each component.} We present detailed results of one-by-one employment of different components in Table~\ref{Ablation Studies}.
As discussed in Table~\ref{tab:challenges of S2I-FSL.}, directly applying prototype refinement results in decreased joint accuracy and $\Delta$ metric, which is caused by the confusion between base and novel classes.
With the help of our proposed meta-training Algorithm~\ref{algorithm:pseudo-unlabeled}, the severe confusion is alleviated and about $11.0\%$/$3.1\%$ absolute performance gains are obtained in $1$/$5$-shot cases. Moreover, model adaptation mechanism that learns discriminative features for novel classes yields another $1.3\%$/$0.4\%$ improvement.

\begin{table}[!htb]
    \scriptsize
    \vspace{-0.1cm}
	\begin{center}
		\begin{tabular}{c | c c c | c c c | c c c}
			\toprule[1pt]
            \multirow{2}*{\textbf{\makecell[c]{Method\\(Semi-supervised)}}} & \multirow{2}*{\textbf{\makecell[c]{Proto\\refine}}} & \multirow{2}*{\textbf{\makecell[c]{Fake\\unlabel}}} & \multirow{2}*{\textbf{\makecell[c]{Model\\adapt}}} & \multicolumn{3}{c|}{\textbf{200+5-way 1-shot}} & \multicolumn{3}{c}{\textbf{200+5-way 5-shot}} \\
            ~ & ~ & ~ & ~ & \textbf{Acc.} & \textbf{$\Delta$}  & \textbf{Acc$_{\text{n/n}}$} & \textbf{Acc.} & \textbf{$\Delta$} & \textbf{Acc$_{\text{n/n}}$} \\
            \midrule 
            Meta-Inc-Baseline & & & & 61.92 & -7.87 & 67.79 & 72.98 & -5.39 & 83.39 \\
            Meta-Inc-Baseline + PR & \checkmark & & & 59.35 & -14.55 & 75.41 & 72.68 & -6.93 & 85.86 \\
            Our method w/o adaptation & \checkmark & \checkmark & & 70.34 & -5.90 & 79.82 & 75.82 & -4.49 & 87.21 \\
            Our full method & \checkmark & \checkmark & \checkmark & \textbf{71.64} & \textbf{-4.97} & \textbf{80.91} & \textbf{76.24} & \textbf{-4.21} & \textbf{87.82} \\
			\bottomrule[1pt]
		\end{tabular}
		
	\end{center}
	\vspace{-1em}
	\caption{Ablation study of each component proposed for S$^2$I-FSL on \emph{tiered}-ImageNet.}
	\label{Ablation Studies}
	\vspace{-0.2cm}
\end{table}

\begin{figure}[!htb]
  \begin{minipage}[c]{0.65\textwidth}
    \centering
    \includegraphics[height=3.28cm]{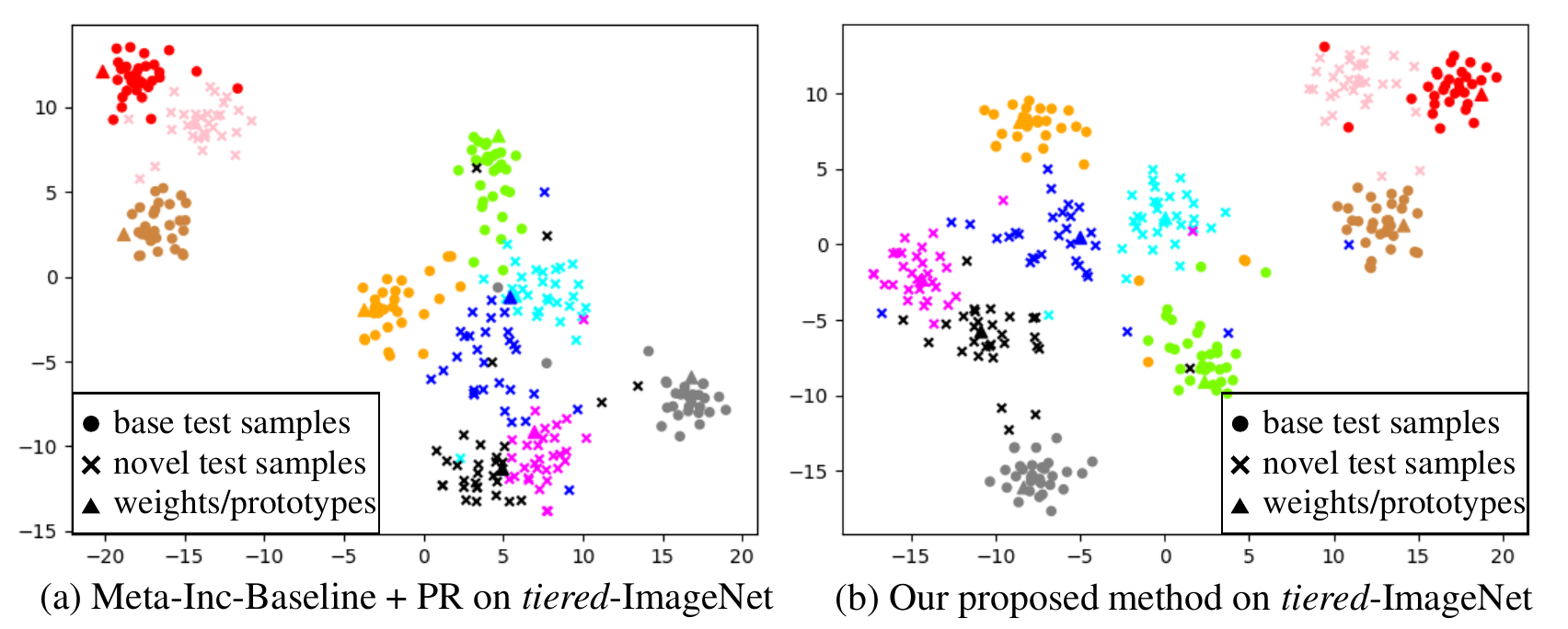}
    \vspace{-1em}
    \caption{T-SNE~\cite{t-SNE} plots of queries and prototypes from a \emph{tiered}-ImageNet test episode with and without our proposed components. Categories are represented by different colors.}
    \label{fig:t-SNE results}
  \end{minipage}
  \hspace{0.005\textwidth}
  \begin{minipage}[c]{0.33\textwidth}
    \centering
    \includegraphics[height=3.28cm]{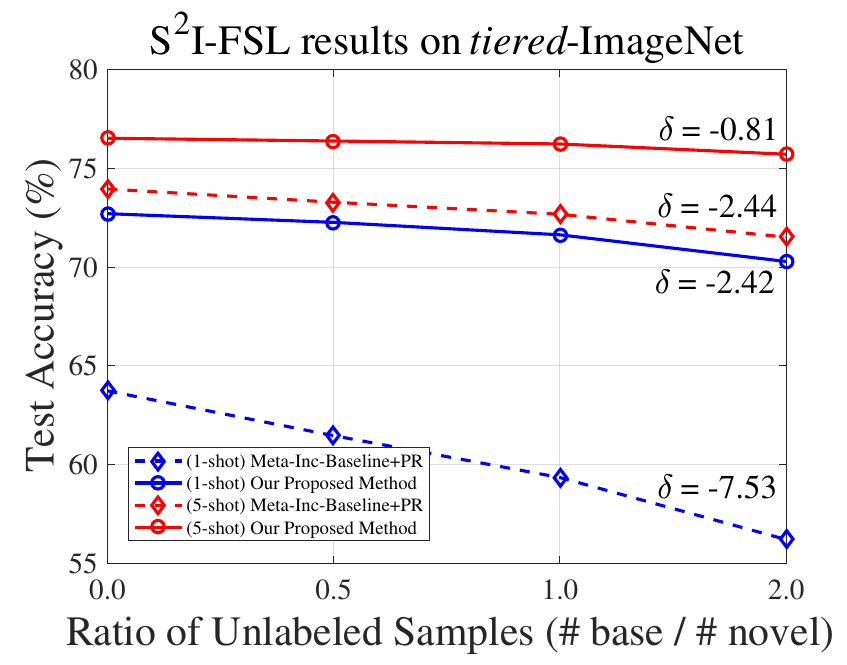}
    \vspace{-1em}
    \caption{Performance of different methods by varying the sample ratio in unlabeled set.}
    \label{fig:Histograms}
  \end{minipage}%
 \vspace{-0.3cm}
\end{figure}

\textbf{Visualization results.} The above observations are also confirmed by t-SNE~\cite{t-SNE} visualizations. Fig.~\ref{fig:t-SNE results} shows a 200+5-way 1-shot semi-supervised incremental test episode on \emph{tiered}-ImageNet without (left) and with (right) our proposed components. For clarity, 5 base classes and 5 novel classes are randomly chosen and features of 30 per-class test samples are considered. The blue novel class prototype (triangle) which was confused with green and orange base classes is more separable from them using Algorithm~\ref{algorithm:pseudo-unlabeled}, and novel class features (blue and cyan, magenta and black) become more distinguishable after model adaptation.

\textbf{Studies on unlabeled set $\mathcal{U}$.} To observe the performance on the second challenge of S$^2$I-FSL mentioned in Section~\ref{sec:intro}, we vary the sample ratio in $\mathcal{U}$ as shown in Fig.~\ref{fig:Histograms}.
The accuracy decreases when fewer novel class samples are included in $\mathcal{U}$, since the more unbalanced samples between base and novel classes, the more challenging the task becomes.
Compared to Meta-Inc-Baseline + PR, our method is more robust to changes in sample ratio, as less degradation ($\delta$) is observed.
More comparison and evaluation details are in Appendix G. 

 \vspace{-0.1cm}

\vspace{-0.4cm}
\section{Conclusion}
\vspace{-0.3cm}
In this paper, a Semi-Supervised Incremental Few-Shot Learning (S$^2$I-FSL) benchmark is proposed to generalize previous semi-supervised FSL and incremental FSL under a more realistic and challenging setting. To solve S$^2$I-FSL, we introduce an efficient meta-baseline for incremental FSL. Based on that, a novel meta-training paradigm and a model adaptation scheme are proposed to fully explore unlabeled data. Extensive experiments on incremental FSL, semi-supervised FSL, and S$^2$I-FSL demonstrate the effectiveness of our method.

\bibliography{template}
\end{document}